%% file: neurips_2025.tex
\documentclass{article}

 \usepackage[preprint]{neurips_2025}

\usepackage[utf8]{inputenc} % allow utf-8 input
\usepackage[T1]{fontenc}    % use 8-bit T1 fonts
\usepackage{hyperref}       % hyperlinks
\usepackage{url}            % simple URL typesetting
\usepackage{booktabs}       % professional-quality tables
\usepackage{amsfonts}       % blackboard math symbols
\usepackage{nicefrac}       % compact symbols for 1/2, etc.
\usepackage{microtype}      % microtypography
\usepackage{xcolor}         % colors
\usepackage{natbib}
\usepackage{graphicx}
\usepackage{algorithm}
\usepackage{algpseudocode}
\usepackage{amsmath}
\usepackage{amssymb}
\usepackage{wrapfig}
\usepackage{bm}
\usepackage{booktabs}
\usepackage{xcolor}
\usepackage{colortbl}
\usepackage[most]{tcolorbox}
\usepackage{multirow} 
\usepackage{enumitem}
\usepackage{siunitx}

\setcitestyle{square}
\setcitestyle{comma}

\hypersetup{
	colorlinks=true,
	citecolor=cyan,
}
\title{Learning to Parallel: Accelerating Diffusion Large Language Models via Learnable Parallel Decoding}

\author{%
  \textbf{Wenrui Bao$^{1}$\thanks{Equal contribution.}~~, Zhiben Chen$^{2}$\footnotemark[1]~~, 
  \textbf{Dan Xu$^{3}$}, 
  \textbf{Yuzhang Shang$^{1}$}\thanks{Corresponding author}}\\
  $^{1}$University of Central Florida,
  $^{2}$Mobi.AI, $^{3}$HKUST \\
\href{https://ims-kdks.github.io/learning-to-parallel/}{\textbf{Project}}~~~~~~\href{https://github.com/ims-kdks/Learning-to-Parallel-Decoding}{\textbf{Code}}
}

\begin{document}

\maketitle

\input{iclr2026/section/0abs}
\input{iclr2026/section/1intro}

\input{iclr2026/section/2related}
\input{iclr2026/section/3method}

\input{iclr2026/section/4exp}

\input{iclr2026/section/5con}

% \begin{ack}
% Use unnumbered first level headings for the acknowledgments. All acknowledgments

% Do {\bf not} include this section in the anonymized submission, only in the final paper. You can use the \texttt{ack} environment provided in the style file to automatically hide this section in the anonymized submission.
% \end{ack}

\bibliography{iclr2026_conference}
\bibliographystyle{iclr2026_conference}
\clearpage

%%%%%%%%%%%%%%%%%%%%%%%%%%%%%%%%%%%%%%%%%%%%%%%%%%%%%%%%%%%%

\appendix

\input{iclr2026/section/6appendix}

\end{document}

%% file: iclr2026/section/0abs.tex
\begin{abstract}
Autoregressive decoding in large language models (LLMs) requires $\mathcal{O}(n)$ sequential steps for $n$ tokens, fundamentally limiting inference throughput. 
Recent diffusion-based LLMs (dLLMs) enable parallel token generation through iterative denoising. However, current parallel decoding strategies rely on fixed, input-agnostic heuristics (e.g., confidence thresholds), which fail to adapt to input-specific characteristics, resulting in suboptimal speed-quality trade-offs across diverse NLP tasks.
In this work, we explore a more flexible and dynamic approach to parallel decoding. We propose \textbf{Learning to Parallel Decode (Learn2PD)}, a framework that trains a lightweight and adaptive filter model to predict, for each token position, whether the current prediction matches the final output. This learned filter approximates an oracle parallel decoding strategy that unmasks tokens only when correctly predicted. Importantly, the filter model is learned in a post-training manner, requiring only a small amount of computation to optimize it (minute-level GPU time). Additionally, we introduce \textbf{End-of-Text Prediction (EoTP)} to detect decoding completion at the end of sequence, avoiding redundant decoding of padding tokens. Experiments on the LLaDA~\citep{nie2025large} benchmark demonstrate that our method achieves up to \textbf{22.58×} speedup without any performance drop, and up to \textbf{57.51×} when combined with KV-Cache.
\end{abstract}

%% file: iclr2026/section/1intro.tex
\section{Introduction}

Large Language Models (LLMs) \citep{LLMSurvey, ziyu-etal-2023-lens, Minaee2024LargeLM} have demonstrated remarkable capabilities across a wide spectrum of natural language processing (NLP) tasks. However, most state-of-the-art LLMs rely on autoregressive (AR) decoding \citep{brown2020language, radford2019language,vaswani2017attention}, which generates output tokens sequentially. Although this approach delivers strong generation quality, it inherently suffers from limited inference efficiency due to its strictly sequential nature \citep{leviathan2023fast, stern2018blockwise}. To overcome this bottleneck, diffusion-based LLMs (dLLMs) \citep{nie2025large, ye2025dream} have been proposed as a compelling alternative by enabling parallel token generation through iterative denoising, potentially achieving sublinear complexity \citep{sohl2015deep, li2022diffusion}.

Diffusion-based LLMs (dLLMs) produce or iteratively refine the entire token sequence via denoising steps rather than predicting tokens one by one, so token-wise predictions at each step can be computed in parallel. Especially, most dLLMs adopt \textit{semi-autoregressive decoding} \citep{arriola2025block}, which divides the target sequence into contiguous blocks and decodes the blocks from left to right. It facilitates token-parallelism by trading a small amount of autoregressive constraint for substantially higher parallel throughput, while still preserving essential left-to-right dependencies. 
To fully unlock these benefits, further development of a parallel decoding strategy that can leverage this approach is needed. 
Current methods employ static heuristics, for example, \textit{confidence-based sampling} \citep{chang2022maskgit} prioritizes the most confident tokens for parallel decoding.
% such as \textit{confidence-based sampling} \citep{chang2022maskgit}, which prioritizes the most confident tokens for parallel decoding.
Although these methods speed up inference, their static decoding strategies lead to poor generation quality. 
% This inherent limitation forces them to compromise on speedup efficiency to maintain generality across diverse tasks. Consequently, they fail to achieve optimal efficiency in real-world scenarios where input complexity varies significantly.

% This inherent limitation makes them difficult to handle diverse and dynamic real-world NLP tasks, as they cannot adapt to the varying difficulty of different inputs—a simple greeting might decode perfectly in one step, while complex reasoning requires iterative refinement.

Targeting this static limitation, we pose an intuitive question: \textit{Instead of relying on a one-rule-fits-all decoding strategy, can we adopt a flexible, case-by-case one for parallel decoding?}
To answer this, we analyzed the model's token-level decoding behavior and found that current models often remask tokens that have already been correctly predicted, leading to unnecessary computational redundancy.
Taking advantage of this finding, we propose that an effective parallel decoding strategy should be capable of eliminating such redundancy.
% \syz{taking advantage of this finding, we propose that a good parallel decoding strategy should be capable of eliminating such redundancy.}
To realize this goal, we first establish an oracle baseline: \textbf{Extremely Greedy Parallel (EGP)}, which unmasks each token immediately upon correct prediction. In the oracle, we use the reference answers to unmask a token when its prediction matches the ground truth.
% \syz{(in the oracle baseline, we use the ground-thuth xxxx, )}.
Our analysis reveals that this oracle can achieve a 15-20× speedup without quality loss, demonstrating substantial potential to improve parallel decoding. However, its dependence on unavailable ground truth makes it infeasible in practice.

% \syz{Our analysis reveals that this oracle achieves 15-20$\times$ speedup with perfect quality, showing there still are some space to squzee the potential of parallel decoding. However, this oracle requires knowing the ground truth information unavailable during inference, which is not possible for real application.}

To approximate this oracle, we propose \textbf{Learning to Parallel Decode (Learn2PD)}, the first learned parallel decoding policy for dLLMs. The framework learns to predict when to finalize a token—that is, when we have sufficient confidence to accept its current prediction.
The key insight is that diffusion models exhibit predictable confidence patterns \citep{song2020denoising, nichol2021improved}: the confidence score for each token can be treated as an informative feature. Fluctuations in these scores capture the model's internal state of acceptance or doubt regarding its predictions.
% tokens with high confidence early in the denoising process tend to the correct prediction, while low-confidence tokens benefit from additional refinement.
Specifically, we train a lightweight filter model $f_\theta$ that predicts whether each token has been correctly generated. The filter model is optimized in the post-training phase, requiring minute-level GPU time for convergence. Once trained, this filter model remains fixed and requires no gradient updates during inference. The filter takes the model’s confidence scores as input and outputs a binary decision for each token to indicate whether it should be remasked.
Surprisingly, a simple two-layer MLP \citep{tolstikhin2021mlpmixer} performs exceptionally well at this task, as the block-level confidence patterns provide sufficient information for accurate convergence prediction, thus eliminating the need for complex architectures or task-specific feature engineering.

% Surprisingly, a simple two-layer MLP \citep{tolstikhin2021mlpmixer} can learn this task very well because the block-level confidence patterns that indicate prediction stability are largely task-agnostic and can be captured by first-order statistics of confidence.
% \syz{Surprisingly, a simple two-layer MLP \citep{tolstikhin2021mlpmixer} can learn this task very well because the block-level confidence patterns contain sufficient signal for accurate convergence prediction, eliminating the need for complex architectures or task-specific feature engineering.}

\begin{figure}[!t]
\centering
% \vspace{-0.2in}
\includegraphics[width=1\textwidth]{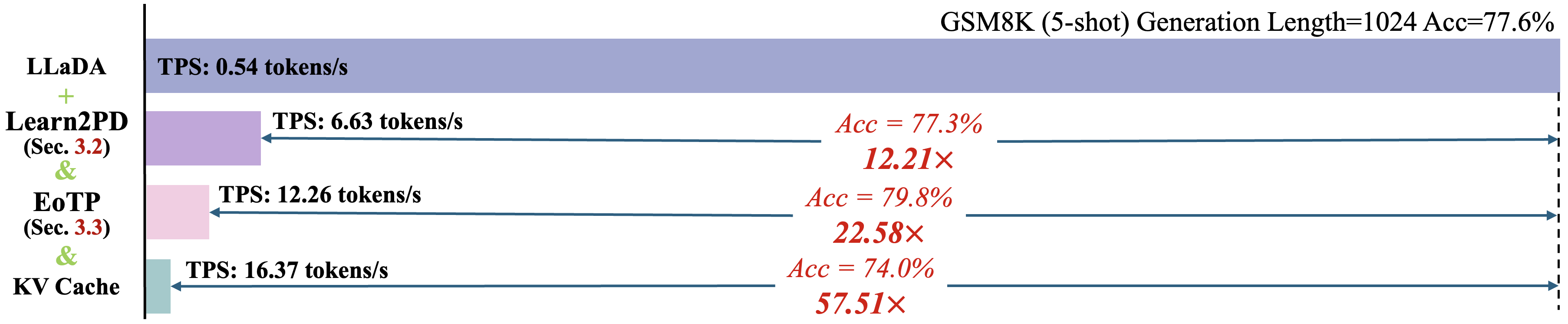}
\vspace{-0.2in}
\caption{\textbf{Effectiveness of our proposed approaches.} We report the throughput and accuracy on GSM8K (5-shot, Generation Length=1024) with LLaDA and our proposed methods under four settings: (1) vanilla decoding, (2) Learn2PD policy, (3) Learn2PD and EoTP mechanism, (4) Learn2PD and EoTP integrated by KV Cache. Our proposed methods, Learn2PD and EoTP, yield a $22.58\times$ speedup over the vanilla baseline while simultaneously preserving the original accuracy. Integration with KV Cache achieves a further improvement in throughput to 16.37 tokens/sec (a 57.51× speedup), with only a minimal loss in accuracy.} 
\vspace{-0.212in}
\label{introduction}
\end{figure}

Another finding from the EGP oracle is that even when the [End-of-Text] token is unmasked, the model continues the decoding process for subsequent tokens. When the generation length is 1024, this inefficiency is responsible for 90\% of the computational waste. To reduce the excessive decoding steps after the [End-of-Text] token, we introduce an \textbf{End-of-Text Prediction (EoTP)} mechanism. EoTP can terminate decoding as soon as the [End-of-Text] token is confidently generated, which avoids redundant computation and further boosts decoding efficiency.

% \syz{Our proposed methods accelerate dLLMs by reducing the redundant decoding operations in dLLMs, and thus do not sacrifice its generation quality. Experiments show our approach achieves \textbf{22.58$\times$} speed-up on the LLaDA without performance drop.}
Our method accelerates dLLMs by eliminating redundant decoding operations, thereby preserving generation quality. Experimental results demonstrate a remarkable \textbf{22.58×} speed-up on LLaDA while fully maintaining its performance.
% Essentially, these techniques reduce redundant decoding operations and achieve substantial acceleration without sacrificing generation quality. As illustrated in Figure~\ref{introduction}, our approach achieves up to a \textbf{22.58$\times$} speed-up on the LLaDA benchmark while maintaining generation performance. 
Importantly, our method is \textbf{orthogonal} to existing optimizations: when combined with KV caching, the speedup compounds to \textbf{57.51×} accompanied by only a slight degradation in accuracy (See Figure \ref{introduction}).
In summary, our contributions are threefold:

\begin{enumerate}[leftmargin=11pt, labelsep=4pt, itemsep=2pt]
    \vspace{-2pt}
    \item We propose a learnable and adaptive framework, \textbf{Learn2PD} that predicts which tokens have been correctly decoded, approximating the oracle Extremely Greedy Parallel Decoding strategy.
    \item We also propose a \textbf{End-of-Text Prediction (EoTP)} mechanism to reduce the unnecessary decoding steps, which significantly boosts inference efficiency.
    \item We extensively evaluate our method on various dLLMs across four representative benchmarks: GSM8K, MATH, HumanEval, and MBPP. Our method consistently achieves order-of-magnitude inference acceleration with negligible accuracy loss. Specifically, our method attains a significant \textbf{22.58×} acceleration without any degradation in accuracy.
\end{enumerate}

%% file: iclr2026/section/2related.tex
\section{Related Work}
% \subsection{Acceleration Methods for Diffusion Models}
% Diffusion models have rapidly advanced visual generation but remain bottlenecked by slow iterative denoising. For classic diffusion models, acceleration methods broadly split into reducing denoising steps and cutting per-step computation. DeepCache \cite{ma2024deepcache} shows that caching high-level upsampling features yields speedups at a small quality cost. As Diffusion Transformers (DiT) become popular, their caching strategies are also coming into focus. FORA\cite{selvaraju2024fora} statically reuses self-attention/MLP outputs at fixed intervals. Learning-to-Cache (L2C) \cite{ma2024learning}learns a timestep-dependent layer router via differentiable interpolation. $\Delta$-DiT \cite{chen2024delta} introduces $\Delta$-Cache and stage-adaptive scheduling. ToCa \cite{zou2024accelerating} selects cached tokens via cheap scores and applies depth/type-aware ratios. DuCa\cite{zou2024acceleratingdu} alternates aggressive (full-block reuse) and conservative (token-wise, FlashAttention-compatible V-caching) steps to bound error accumulation. Additionally, TGATE \cite{liu2025faster} temporally gates attention by observing that cross-attention converges and is redundant in late steps, enabling caching/reuse of cross-attention and sparse early self-attention. EB‑Cache \cite{zou2025exposure} uses a cache table to control noise scaling and achieves acceleration with very low quality loss.

\subsection{Diffusion-based large language models}
The integration of diffusion models with large language models (LLMs) is an emerging and promising direction in generative AI. Early work adapted continuous diffusion to discrete data domains \citep{sohl2015deep, hoogeboom2021argmax}, leading to D3PM \citep{austin2021structured}, which introduced a Markov chain-based framework for discrete noise injection and denoising trained via ELBO maximization. This was extended to continuous time by CTMC \citep{campbell2022continuous}. In parallel, SEDD \citep{lou2023discrete} learned the reverse process by modeling the ratio of marginal probabilities using a denoising score entropy objective, while Masked Diffusion Models such as MDLM \citep{shi2024simplified, sahoo2024simple, zheng2024masked} and RADD \citet{ou2025your} provided further theoretical simplifications and formalized connections between parameterizations. A key breakthrough has been the incorporation of diffusion into existing LLM architectures: Diffusion-NAT \citep{Zhou2023DiffusionNATSD} aligned the denoising process with non-autoregressive decoding, enabling high-speed generation, while models like LLaDA \citep{nie2025large}, DiffuLLaMA \citep{gong2025scaling}, and Dream \citep{ye2025dream} successfully scaled diffusion-based decoding to billion-parameter models, significantly improving inference efficiency without compromising output quality.
% These developments demonstrate the strong potential of diffusion models in language generation.

\subsection{Accelerate Diffusion-based large language models}
Followed by mature diffusion large language models, their acceleration methods are also under development. Concretely, dllm-Cache \citep{liu2025dllm} proposes a training-free, adaptive caching framework that performs long-interval prompt caching and short-interval, value-similarity–guided partial response updates. Fast-dLLM  \citep{wu2025fast} introduces block-wise approximate KV caching and a confidence-aware parallel decoding rule that only decodes tokens whose marginal confidence exceeds a threshold. \citet{hu2025accelerating} propose FreeCache to approximate KV states by reusing stable prompt/block activations across steps. They also introduce Guided Diffusion to decide which tokens to unmask each step without retraining. SlowFast-Sampling \citep{wei2025accelerating} proposes a dynamic two-stage sampler that alternates a cautious exploratory phase with a fast phase that aggressively decodes high-confidence tokens within that span. 
Prophet \citep{li2025diffusion} monitors the top-2 logit gap and commits all remaining tokens in one shot via early-commit decoding once it is sufficiently confident.
These acceleration methods are often static and lack flexibility. 
APD \citep{israel2025acceleratingdiffusionllmsadaptive} employs a small autoregressive model to dynamically control the number of tokens generated in parallel. But they fix the dLLM decoding process to proceed left-to-right, which prevents them from fully exploiting the parallelization advantages of dLLMs.
To address this, we propose \textbf{Learn2PD}, a novel dynamic and parallel decoding method that achieves more efficient inference acceleration by reducing the unnecessary and repetitive decoding steps. Moreover, we also introduce \textbf{EoTP} to avoid redundant decoding when the answer does not span the full generation length.

%% file: iclr2026/section/3method.tex
\section{Methodology}
In this section, we present Learn2PD, a learned approach to accelerate diffusion language model inference through adaptive parallel decoding. We begin by reviewing the fundamentals of diffusion language models and their current parallel decoding strategies (Section~\ref{dllm}). Through empirical analysis, we reveal a critical inefficiency: existing methods unnecessarily remask a significant proportion of correctly predicted tokens, leading to redundant computation (Section~\ref{sec:unnecessary}). This observation motivates our core contribution—training a lightweight filter model to predict token stability and approximate an oracle parallel decoding strategy (Section~\ref{sec:learn2pd}). Finally, we introduce an early-stopping mechanism to further eliminate padding token overhead (Section~\ref{sec:eotp})
% \syz{In this section, we present Learn2PD, a learned approach to accelerate diffusion language model nference through adaptive parallel decoding. We begin by reviewing the fundamentals of diffusion language models and their current parallel decoding strategies (Section~\ref{dllm}). Through empirical analysis, we reveal a critical inefficiency: existing methods unnecessarily remask a significant proportion of correctly predicted tokens, leading to redundant computation (Section~\ref{sec:unnecessary}). This observation motivates our core contribution—training a lightweight filter model to predict token stability and approximate an oracle parallel decoding strategy (Section~\ref{sec:learn2pd}). We then introduce an early-stopping mechanism to further eliminate padding token overhead (Section~\ref{sec:eotp}). Finally, we present our complete algorithms for both training and inference (Section~\ref{sec:algorithm}).}

\subsection{Preliminary}
\subsubsection{Diffusion Large Language Models}
\label{dllm}
\paragraph{Forward Process.}
Given an input sentence $\bm{x}_0 \in \{0,1,\ldots,V-1\}^L$ and a noise level $t \in [0,1]$, where $V$ and $L$ represent the vocabulary size and sentence length. The forward process randomly and independently masks out tokens through the following Markov chain:
\begin{equation}
q_{t\mid 0}(\bm{x}_t \mid \bm{x}_0)
= \prod_{i=0}^{L-1}
\Big[
(1-t)\,\mathbf{1}\{\bm{x}_t^{i}=\bm{x}_0^{i}\}
+ \,t\cdot\mathbf{1}\{\bm{x}_t^{i}=m\}
\Big]
\end{equation}
where $x^{i}$ denotes the $i$-th element of $\bm{x}$, $m$ denotes the mask token \citep{devlin2019bert}, $\bm{x}_t$ denotes the noisy data at time $t$, and $q_0(\cdot)$ is the data distribution $p_{\text{data}}(\cdot)$ .

\paragraph{Reverse process.}
The reverse process iteratively recovers masked tokens by predicting data distribution from a masked sequence. Transitioning from corruption level $t$ to an earlier level $s$, where $0\leq s < t \leq 1$ can be approximated as
\begin{equation}
q_{s\mid t}(\bm{x}_s \mid \bm{x}_t) \;=\; \prod_{i=0}^{L-1} q_{s\mid t}\!\left(\bm{x}_s^{i} \mid \bm{x}_t\right), 
q_{s\mid t}\!\left(\bm{x}_s^{i} \mid \bm{x}_t\right) \;=\;
\begin{cases}
1, & \bm{x}_t^{i} \neq m,\; \bm{x}_s^{i} = \bm{x}_t^{i},\\[4pt]
\dfrac{s}{t}, & \bm{x}_t^{i} = m,\; \bm{x}_s^{i} = m,\\[8pt]
\dfrac{t-s}{t}\; q_{0\mid t}\!\left(\bm{x}_s^{i} \mid \bm{x}_t\right), & \bm{x}_t^{i} = m,\; \bm{x}_s^{i} \neq m,
\end{cases}
\label{eq:reverse}
\end{equation}
where $m$ represent the \texttt{[MASK]} and $q_{0\mid t}(\cdot)$ is the data prediction distribution by the model \citep{ho2020denoising}. 
Given a prompt $\bm{c} = (c_1,...,c_M)$, the response $y$ is generated in $K$ discrete steps. In each step $k$, a mask predictor $p_\theta$ takes $\mathbf{y}^{(k)}$ as input and predicts the distribution of sequence. The estimate of the sequence $\hat{\bm{y}}^{(0)}$ is generated via greedy decoding:
\begin{equation}
    \bm{\hat{y}}^{(0)} = \arg\max_{\bm{y} \in \mathcal{T}} \, P_{\theta}\bigl(\bm{y} \mid \mathbf{c}, \mathbf{y}^{(k)}\bigr) = \arg\max_{\bm{y} \in \mathcal{T}} \, p_{\theta}\bigl(\mathbf{c}, \mathbf{y}^{(k)}; \theta \bigr)
\end{equation}

\begin{wrapfigure}{r}{0.52\textwidth}
\vspace{-1.2em}
% \vspace{-5em}
\centering
\includegraphics[width=1\linewidth]{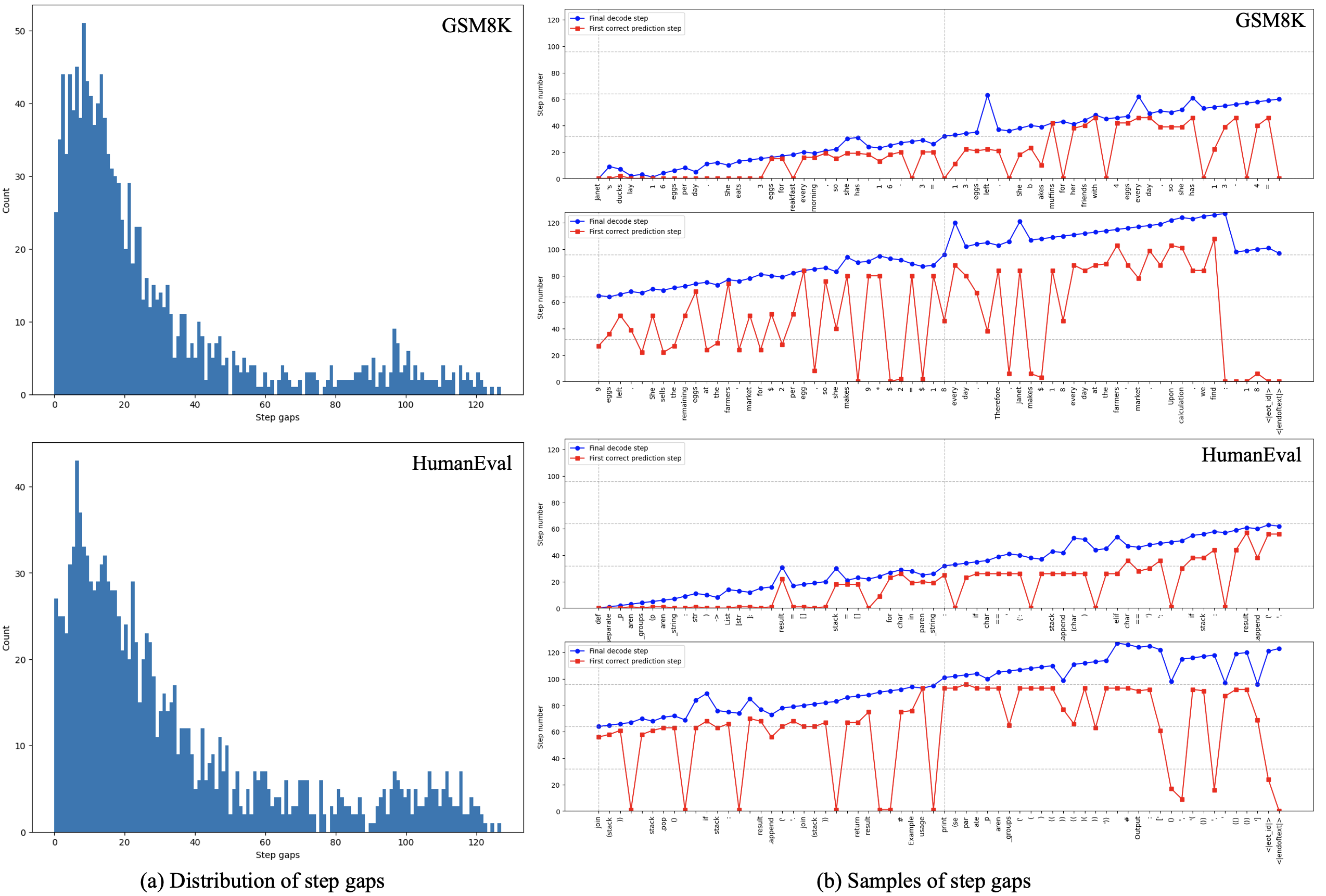}
\vspace{-0.2in}
\caption{\textbf{The unnecessary and repetitive decoding steps in different datasets: GSM8K and HumanEval.} (a) Distributions of gaps. These two histograms show the distribution of step gaps for each token between the decoding step and the step with the first correct prediction. (b) Samples of gaps. The red line means the first correct prediction step, and the blue line means the actual decoding step.}
\label{stepgap}
\vspace{-2em}
\end{wrapfigure}

\paragraph{Low-Confidence Remasking.} To improve the sample quality, the unmasking tokens with low confidence would be remasked. This approach follows a common practice in non-autoregressive generation for improving output fidelity \citep{ghazvininejad2019mask}. For each position $i$, the model predicts $\hat{y_0}^{(k)}$ and computes its confidence $c_i$, which is given by:
\begin{equation}
    c_i = P_{\theta}\bigl(\hat{y_{0,i}}^{(k)} \mid \mathbf{c}, \mathbf{y}^{(k)}\bigr)
\end{equation}
The tokens corresponding to the $n$ lowest confidence would be set to \texttt{[MASK]} again, where $n$ is calculated by the noise level $t$.

% \paragraph{Semi-autoregressive Remasking.} To improve the generation quality of large language diffusion models, the sequence to be generated will be divided into several blocks, which will be generated from left to right one by one.

\subsubsection{Unnecessary repetitive decoding}
\label{sec:unnecessary}

Building on the iterative inference process displayed in Section \ref{dllm}, we investigate the unnecessary and repetitive decoding conditions in diffusion-based large language models. We conducted experimental analyses with LLaDA-8B-Instruct\citep{nie2025large} on two widely used datasets: GSM8K \citep{cobbe2021gsm8k} and HumanEval \citep{chen2021codex}. 
We choose LLaDA as our base model due to its state-of-the-art performance and availability of pre-trained checkpoints across multiple scales.
Specifically, we measured the amount of unnecessary and repetitive decoding, which is defined as the number of times the model continues to decode a token after that token has first matched the reference answer. In this paper, we refer to the answer produced by LLaDA under the standard generation process as the \textbf{reference answer}.\footnote{For all analyses in this section, we set LLaDA's Generation Length at 128 and Block Size at 32.}

% For all analyses, we set LLaDA's Generation Length at 128 and Block Size at 32.

\paragraph{Analysis of unnecessary repetition.} As illustrated in Figure \ref{stepgap} (a), we tally up the distribution of the step gaps between the decoding step and the step with the first correct prediction. For each dataset, we randomly sample 10 questions to conduct the experiment. We find that most of the tokens still need to be decoded more than 10 times, even though they are already correct. And Figure \ref{stepgap} (b) shows one sample from each dataset. The red line means the first correct prediction step, and the blue line means the actual decoding step. It is clear that the model performs many unnecessary decoding steps before unmasking the tokens.

\begin{figure}[!t]
\centering
\includegraphics[width=1.0\textwidth]{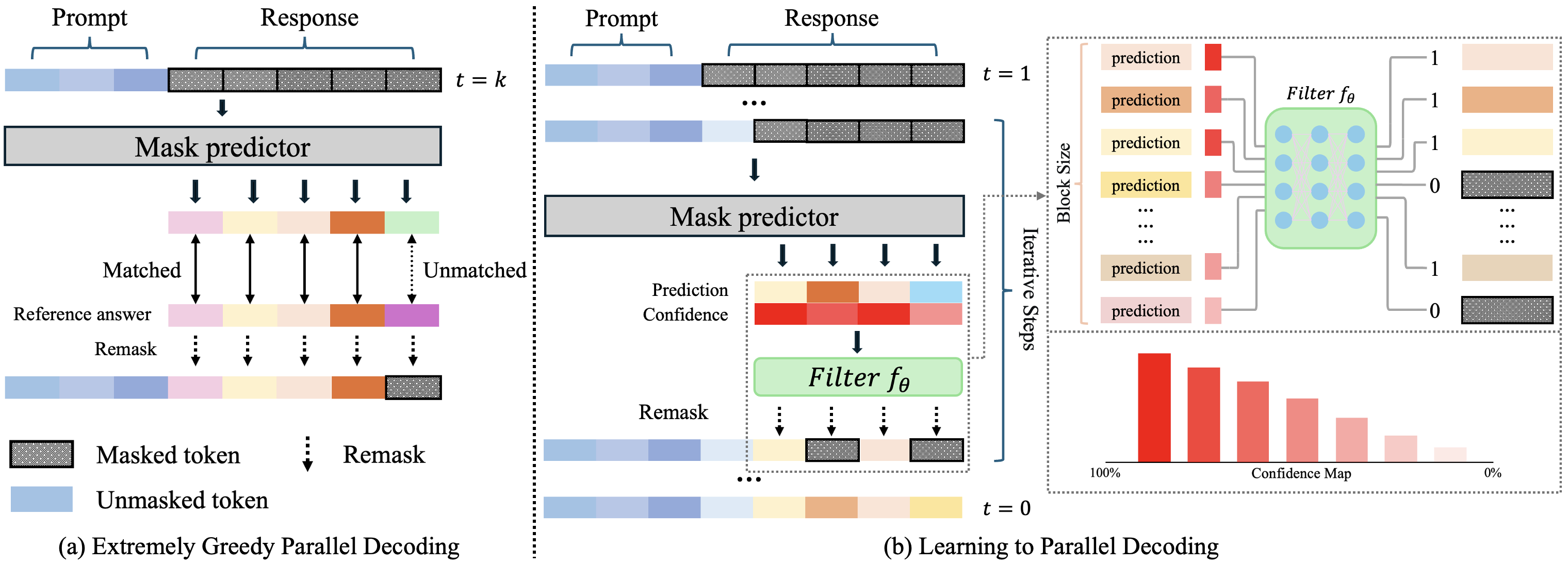}
\vspace{-0.2in}
\caption{\textbf{A Conceptual Overview of pipeline and method.} (a) Extremely Greedy Parallel (EGP). This strategy compares the predicted tokens with the reference answer and only remasks the tokens that do not match in these comparisons. (b) Learning to Parallel Decoding (Learn2PD). During the inference process, after the model generates predictions and confidences for each token, the confidence of each token is fed into a filter model $f_\theta$ to determine which tokens need to be remasked. This determination then guides the subsequent remasking procedure.}
\vspace{-0.2in}
\label{fig:overview}
\end{figure}

\subsubsection{Expected Inference Process: Extremely Greedy Parallel (EGP)}

Based on the above findings, we observe that a large portion of tokens are remasked as \texttt{[MASK]} and decoded multiple times even after they have already been decoded to the reference answer. Motivated by this, we define the \textbf{Extremely Greedy Parallel (EGP)} oracle (Figure \ref{fig:overview}a) as:
at each step k, unmask token $i$ if and only if $M(x^k)_i = y_i$,
where $y_i$ is the reference answer for token $i$. This oracle achieves optimal
speedup by never remasking correct predictions.
% we propose a strategy called “\textbf{Extremely Greedy Parallel Decoding}”. In this strategy, whenever a token is first decoded to the correct answer, we set it to unmasked. We illustrate details of our proposed policy in Figure \ref{fig:overview}. By following this approach, the model can produce the same final answer without any redundant decoding. Here is the new remask process: For each token $t$ and discrete step $k$, we have:
% \begin{equation}
%     t^{(k)} \;=\;
% \begin{cases}
% t^{(k)}_{predict}, \quad t^{(k)}_{predict}=t_{reference},\\[4pt]
% \text{[MASK]}, \quad otherwise.
% \end{cases}
% \end{equation}
% where $t_{reference}$ represents the reference answer for token $t$. 

\begin{wrapfigure}{r}{0.48\textwidth}
\centering
\vspace{-0.2in}
\includegraphics[width=\linewidth]{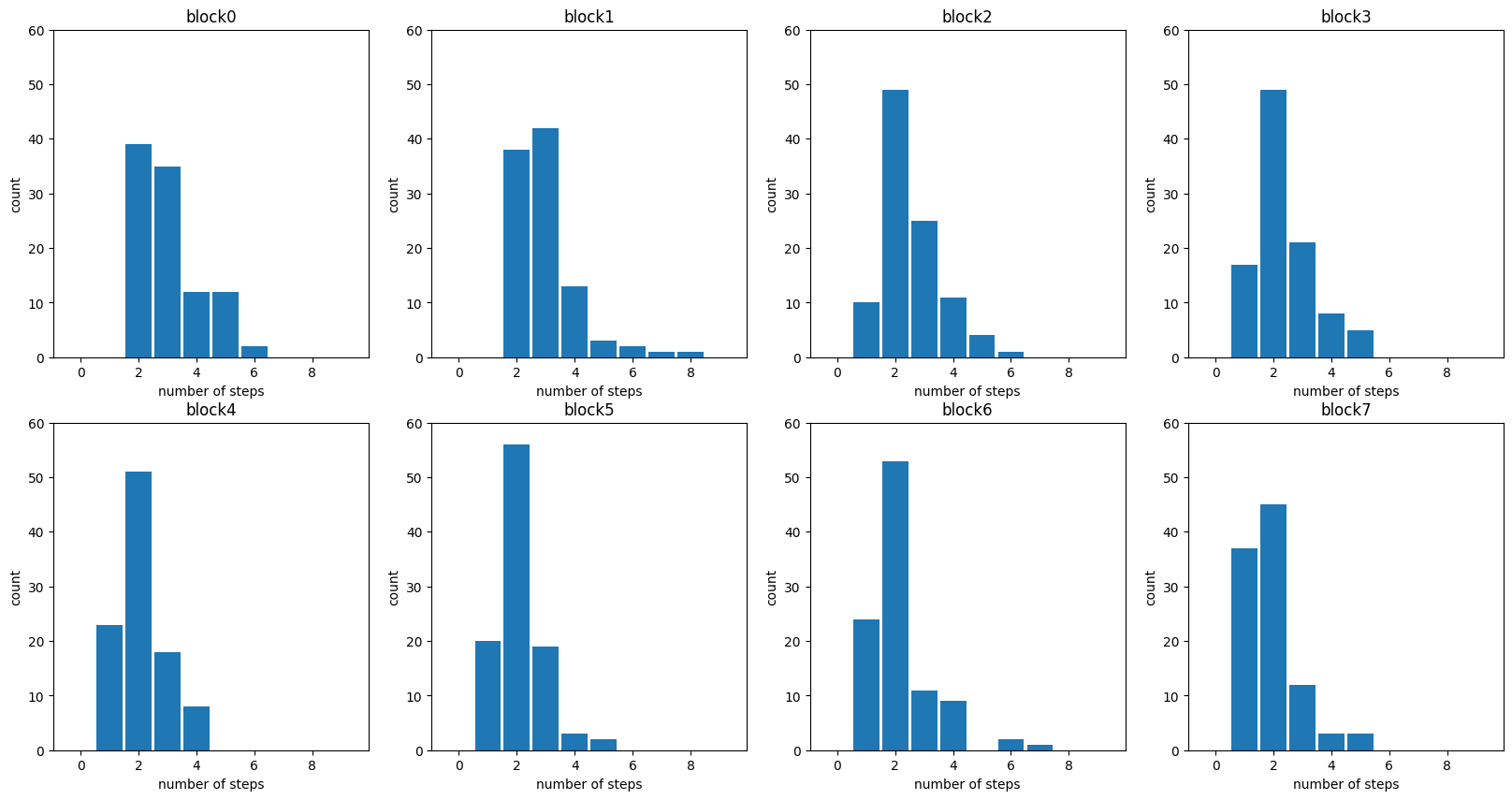}
\vspace{-0.25in}
\caption{\textbf{Distribution of decoding steps per block with Extremely Greedy Parallel (EGP) strategy.} Histograms illustrate the number of decoding steps performed in each block when using our strategy with LLaDA-8B-Instruct on GSM8K based on 100 samples.}
\label{potential}
\vspace{-2em}
\end{wrapfigure}

\paragraph{Acceleration Potential.} To evaluate the efficiency of our strategy, we compared the number of decoding steps required per block for LLaDA-8B-Instruct on the GSM8K dataset under the \textbf{Extremely Greedy Parallel} policy versus the standard decoding regime. Similarly, we fix the Generation Length to 256 and the Block Size to 32.

As shown in Figure \ref{potential}, the results are striking. Our strategy achieves a median of 2 decodings per block while maintaining the same accuracy. In contrast, LLaDA with the vanilla setting requires 32 decodings per block. This demonstrates a substantial opportunity for efficiency gains, without compromising output quality.

% As shown in Figure \ref{potential}, the results are striking. When using our strategy, the median number of block decodings is 2 across all blocks with the same accuracy. In contrast, LLaDA with vanilla setting performs 32 decodings per block. This indicates that there is significant latent space for efficiency gains without sacrificing output length or structure.

\subsection{Learning to Parallel Decoding}
\label{sec:learn2pd}
Although our Extremely Greedy Parallel strategy performs well, this oracle requires ground truth tokens that are unavailable during inference. To address this, we propose a novel approach: \textbf{Learning to Parallel Decoding (Learn2PD)} (Figure \ref{fig:overview}b). Our goal is to simulate the EGP strategy after each decoding step to select tokens and decide whether to remask.  We can reformulate this as an optimization problem by using Binary Cross-Entropy Loss (BCELoss) \citep{de2005tutorial}:
\begin{equation}
\arg\min\; -\frac{1}{m}\sum_{i=1}^m\Big[y_i\log p_i + (1-y_i)\log(1-p_i)\Big]
\end{equation}
where $y_i$ indicates whether token $t_i$ should be remasked under the EGP strategy: 0 means it should be remasked and 1 means it can be unmasked. During the inference process, a threshold $\tau$ is applied to discretize $p_i$ into either 0 or 1. The $p_i$ is generated from which we called the filter model $f_\theta$. In this algorithm, the only trained parameters are $\theta$. Therefore, the parameters of the diffusion large language model remain unchanged. Then, the training loss should be:
\begin{equation}
    \mathcal{L}_{BCE} = -\frac{1}{m}\sum_{i=1}^m\Big[y_i\log \sigma(z_i) + (1-y_i)\log(1-\sigma(z_i))\Big]
\end{equation}
where $z_i$ is the output of the filter model (logit).
% \subsubsection{Algorithms}
% \label{sec:algorithm}
We show the algorithms for training and inference in Algorithm \ref{training} and Algorithm \ref{inference}. The filter $f_\theta$ takes the confidence of the prediction as input and returns the logits $z$ to indicate the probability of no remask. And in order to ensure $z$ remains in the range $[0,1]$, we apply a sigmoid function on $z$ before it is passed to the dLLMs. 
Critically, the filter model $f_\theta$ adds negligible overhead during inference. Our experimental results in Section \ref{sec:exp} quantitatively demonstrate that the achieved speedup vastly outweighs this minimal overhead.

\begin{minipage}{0.48\textwidth}    
    \vspace{-0.1in}
    \begin{algorithm}[H]
    \scriptsize
    \caption{Training}
    \begin{algorithmic}[1]
    \Require Diffusion large language model $M$, filter model $f_\theta$, prompt set $x_\text{prompt}$, reference answer set $x_\text{reference}$, generation length $L_\text{gen}$, learning rate $\eta$, block size $s$
    \Repeat
        \State $x_i \in x_\text{prompt}$, $r_i \in x_\text{reference}$, $l_i = length(x_i)$
        \State $X \gets \text{concat}(x_i, [\text{MASK}]^{L_\text{gen}})$
        \For{$b = 0,...,\frac{L_\text{gen}}{s}-1$}
            \State $\mathcal{M} \gets \{1,2,...,s\}$
            \While{$\mathcal{M} \neq \emptyset$ }
                \State $conf_t, \; pre_t= M(X)$
                \If{$pre_{t,j}=r_{i,j}$}
                    \State $\hat{y_j} \gets 1$, $\mathcal{M} \gets \mathcal{M} \setminus \{j\}$
                    \State $X_{l_i+b\cdot s+j} \gets pre_{t,j}$
                \Else
                    \State $\hat{y_j} \gets 0$ 
                \EndIf
                \State $\mathcal{L} \gets \text{BCELoss}(\hat{y}, f_\theta(conf_t))$
                \State $\theta \leftarrow \theta - \eta \cdot \nabla_\theta L$
            \EndWhile
        \EndFor
    \Until{converged}
    \end{algorithmic}
    \label{training}
    \end{algorithm}

\end{minipage}
\hfill
\begin{minipage}{0.48\textwidth}
    \vspace{-0.1in}
    \vspace{-3.1em}
    \begin{algorithm}[H]
    \scriptsize
    \caption{Inference}
    \begin{algorithmic}[1]
    \Require Diffusion large language model $M$, filter model $f_\theta$, prompt set $x_\text{prompt}$, generation length $L_\text{gen}$, block size $s$, filter threshold $\tau$
    \For{each $x_i \in x_\text{prompt}$}
        \State $l_i = \text{length}(x_i)$, $X \gets \text{concat}(x_i, [\text{MASK}]^{L_\text{gen}})$
        \For{$b = 0,...,\frac{L_\text{gen}}{s}-1$}
            \State $\mathcal{M} \gets \{1,2,...,s\}$
            \While{$\mathcal{M} \neq \emptyset$ }
                \State $\text{conf}_t, \; \text{pre}_t= M(X)$, \; $\text{logit}_t = f_\theta(\text{conf}_t)$
                \If{$\text{logit}_{t,j} > \tau$}
                    \State $\mathcal{M} \gets \mathcal{M} \setminus \{j\}$, $X_{i+b\cdot s+j} \gets \text{pre}_{t,j}$
                    % \State $X_{i+b\cdot s+j} \gets pre_{t,j}$
                \EndIf
            \EndWhile
        \EndFor
        \State $\text{response}_i = X_{l_i:l_i+L_\text{gen}-1}$
    \EndFor
    \State \textbf{return} response
    \end{algorithmic}
    \label{inference}
    \end{algorithm}

\end{minipage}

\subsection{End-of-Text Prediction}
\label{sec:eotp}
\begin{wrapfigure}{r}{0.5\textwidth}
    \vspace{-5.6em}
    \centering
    \includegraphics[width=0.9\linewidth]{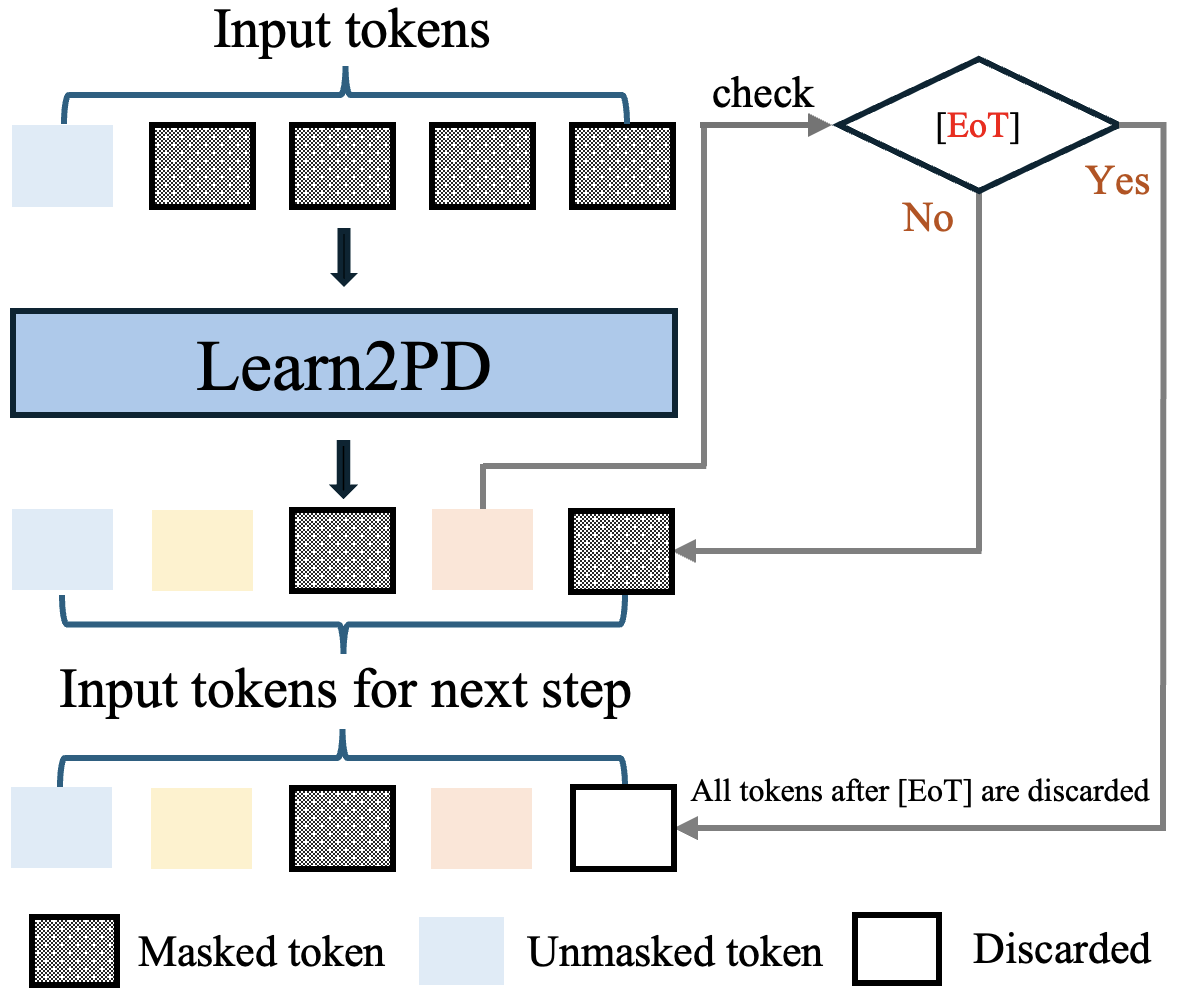} % 图片宽度为预留宽度的90%
    % \vspace{-0.1in}
    \caption{\textbf{Schematic of the End-of-Text Prediction Policy.} During the inference process, upon detection of an \texttt{[EoT]} token in a decoded block, all subsequent tokens are discarded.}
    \label{fig:wrapdemo}
    \vspace{-0.5em}
\end{wrapfigure}

Besides the methods mentioned earlier, we observed that when the generation length of a diffusion large language model is increased to 1024, the generation time rises significantly for the same question compared to a length of 256, even though the final answer length remains unchanged. According to the analysis of the generated output, we find that the extra length is filled with the \texttt{[EoT]} token, and the additional decoding time is spent repeatedly decoding the \texttt{[EoT]} token. Based on this, we propose the \textbf{End-of-Text Prediction (EoTP)} approach: at each step, when a \texttt{[EoT]} token is unmasked, discard tokens in subsequent positions. The shortened sequence is then used as the input of next step. Therefore, we update the inference process to handle the long-generation-length challenge in Appendix \ref{appendix:update_algorithm}. We illustrate the details in Figure \ref{fig:wrapdemo}. Our analysis shows that 89.59\% of computational cost comes from decoding padding tokens after \texttt{[EoT]}. EoTP yields substantial computational savings by dynamically reducing the effective input length throughout the diffusion process. The experiments and relevant analysis are in Appendix \ref{appendix:eotp_analysis}.
% \syz{Our analysis shows that xxx\% of computational cost comes from
% decoding padding tokens after EOS. EoTP eliminates this overhead by
% detecting sequence completion when all non-\texttt{[MASK]} positions have
% confidence > 0.99 for two consecutive steps.}

%% file: iclr2026/section/4exp.tex
\sisetup{
  round-mode = places,      % 四舍五入到指定小数位
  round-precision = 2,      % 保留2位小数
  table-number-alignment = left, % 表格数字对齐方式
  table-format = 2.2,       % 默认格式：2位整数，2位小数
}
{
\setlength{\extrarowheight}{2.3pt}
\begin{table}[!t]
    \centering
    \footnotesize
    \caption{Benchmark results on the LLaDA-8B-Instruct suite. Each method was evaluated using two generation lengths (256 and 1024) across four datasets. Performance is measured using three metrics: TPS (tokens/sec), speedup, and accuracy score. The highest throughput and speedup values for each configuration are highlighted in bold.}
    \begin{tabular}{c|c|c|S|c|c}
        \toprule
        \multirow{2}{*}{\textbf{Task}} & \multirow{2}{*}{\textbf{Methods}} & \multirow{2}{*}{\textbf{Gen Length}} & \multicolumn{2}{c}{\textbf{Inference Efficiency}} & \textbf{Performance} \\
        % \cmidrule(lr){4-6}
        \cline{4-6}
        & & &\textbf{TPS$\uparrow$} & \textbf{Speed (TPS)$\uparrow$} & \textbf{Score} \\
        % \midrule
        \cline{1-6}
        \multirow{6}{*}{\shortstack{GSM8K \\ (5-shot)}} & \multirow{2}{*}{LLaDA-8B-Instruct} & 256 & 3.41 & 1.00$\times$ & 78.70 \\
        & & 1024 & 0.54 & 1.00$\times$ & 77.60 \\
        % \cmidrule(lr){2-6}
        \cline{2-6}
        & \multirow{2}{*}{+ Learn2PD} & 256 & ${14.07}_{\textcolor{green!50!black}{+10.66}}$ & 4.13${ \times}$ & 78.62 \\
        & & 1024 & ${6.63}_{\textcolor{green!50!black}{+6.09}}$ & 12.21${ \times}$ & 77.26 \\
        % \cmidrule(lr){2-6}
        \cline{2-6}
        & \multirow{2}{*}{Learn2PD + EoTP} & 256 & ${14.35}_{\textcolor{green!50!black}{+10.94}}$ & 4.21${ \times}$ & 78.62 \\
        & & 1024 & ${12.26}_{\textcolor{green!50!black}{+11.72}}$ &\textbf{22.58}${ \times}$ & 79.83 \\
        % \midrule
        \cline{1-6}
        \multirow{6}{*}{\shortstack{Math\\(4-shot)}} & \multirow{2}{*}{LLaDA-8B-Instruct} & 256 & 4.70 & 1.00$\times$ & 32.90 \\
        & & 1024 & 1.70 & 1.00$\times$ & 35.21 \\
        % \cmidrule(lr){2-6}
        \cline{2-6}
        & \multirow{2}{*}{+ Learn2PD} & 256 & ${15.16}_{\textcolor{green!50!black}{+10.46}}$ & 3.21${ \times}$ & 32.22 \\
        & & 1024 & ${10.98}_{\textcolor{green!50!black}{+9.28}}$ &  6.45${\times}$ & 34.01 \\
        % \cmidrule(lr){2-6}
        \cline{2-6}
        & \multirow{2}{*}{Learn2PD + EoTP} & 256 & ${15.21}_{\textcolor{green!50!black}{+10.51}}$ & 3.23${ \times}$ & 31.40 \\
        & & 1024 & ${12.27}_{\textcolor{green!50!black}{+10.57}}$ & \textbf{7.22}${ \times}$ & 34.60 \\
        % \midrule
        \cline{1-6}
        \multirow{6}{*}{\shortstack{HumanEval \\(0-shot)}} & \multirow{2}{*}{LLaDA-8B-Instruct} & 256 & 3.33 & 1.00$\times$ & 39.63 \\
        & & 1024 & 0.53 & 1.00$\times$ & 37.21 \\
        % \cmidrule(lr){2-6}
        \cline{2-6}
        & \multirow{2}{*}{+ Learn2PD} & 256 & ${11.66}_{\textcolor{green!50!black}{+8.33}}$ & 3.5${\times}$ & 38.41 \\
        & & 1024 & ${4.63}_{\textcolor{green!50!black}{+4.10}}$ & 8.78${ \times}$ & 37.84 \\
        % \cmidrule(lr){2-6}
        \cline{2-6}
        & \multirow{2}{*}{Learn2PD + EoTP} & 256 & ${11.88}_{\textcolor{green!50!black}{+8.55}}$ & 3.57${\times}$ & 38.41 \\
        & & 1024 & ${6.63}_{\textcolor{green!50!black}{+6.10}}$ & ${\textbf{12.55} \times}$ & 35.98 \\
        % \midrule
        \cline{1-6}
        \multirow{6}{*}{\shortstack{MBPP \\(3-shot)}} & \multirow{2}{*}{LLaDA-8B-Instruct} & 256 & 3.14 & 1.00$\times$ & 31.22 \\
        & & 1024 & 0.58 & 1.00$\times$ & 10.61 \\
        % \cmidrule(lr){2-6}
        \cline{2-6}
        & \multirow{2}{*}{+ Learn2PD} & 256 & ${14.96}_{\textcolor{green!50!black}{+11.82}}$ & 4.77${ \times}$ & 30.84 \\
        & & 1024 & ${6.96}_{\textcolor{green!50!black}{+6.38}}$ & 12.08${ \times}$ & 10.04 \\
        % \cmidrule(lr){2-6}
        \cline{2-6}
        & \multirow{2}{*}{Learn2PD + EoTP} & 256 & ${15.88}_{\textcolor{green!50!black}{+12.74}}$ & 5.06${ \times}$ & 31.03 \\
        & & 1024 & ${9.89}_{\textcolor{green!50!black}{+9.31}}$ & \textbf{17.16}${ \times}$ & 11.02 \\
        \bottomrule
    \end{tabular}
    \label{results}
\end{table}
}
\section{Experiment}
\label{sec:exp}
\subsection{Experimental Settings}
\paragraph{Models and Datasets.} We implement our methods on the representative dLLM: LLaDA-8B-Instruct \citep{nie2025large} to measure the acceleration of the inference process across various benchmarks. To ensure the broad applicability of the methods, we conducted experiments on four datasets covering three different types of problems, which are GSM8K\citep{cobbe2021gsm8k}, Math \citep{lewkowycz2022solving}, HumanEval\citep{chen2021codex}, and MBPP \citep{austin2021program}. All experiments are conducted on 4 NVIDIA A6000 GPUs.

\paragraph{Filter Model $f_\theta$ Training.} To train a filter model that can be applied to a wide range of tasks, we selected 40 samples from each of the 66 types of questions in the FLAN dataset, resulting in a total of 2,640 samples for training. In this experiment, we used the simplest two-layer MLP as our filter model. Since the dLLMs remains frozen and only $f_\theta$ is trained, the number of trainable parameters is extremely limited. For example, for an LLaDA with a block size of 32, the total number of trainable parameters is only 2,112. We trained $f_\theta$ for 5,000 epochs until the model converged. The learning rate is set to 0.001, and the AdamW optimizer is used to optimize $f_\theta$.

Our training process consists of two stages.  In the first stage, samples are collected by following an Extremely Greedy Parallel policy, recording the confidence scores and token selections at each step during parallel decoding. This data is then used in the second stage to train a filter model $f_\theta$. The data collection in the first stage was conducted on 4 NVIDIA RTX A6000 GPUs and took approximately three hours. The subsequent training of the filter model in the second stage was deployed on a T4 GPU and required only 6 minutes. The details of training are in Appendix \ref{appendix:training_curve}.

\paragraph{Evaluation.} We evaluate the inference acceleration and generation quality of \textbf{Learn2PD} and \textbf{EoTP} methods by using quantitive metrics. The inference speed is quantified with Tokens Per Second (TPS), indicating the average number of tokens generated per second. And the generation quality is measured in task-specific metrics, such as accuracy for GSM8K, showing the model's performance with acceleration methods. In addition to this, we set the Generation Length to 256 \& 1024 and the Block Size to 32.

\subsection{Main Results}
We present the inference performance and efficiency profits for Learn2PD and EoTP on the LLaDA-8B-Instruct across four benchmarks, as shown in Table \ref{results}.

In summary, Learn2PD significantly enhances inference efficiency across all tasks. Compared to the baseline model, our optimal method typically achieves a 3 to 4 times speedup at a generation length of 256 and a 6 to 12 times speedup at a generation length of 1024.   When EoTP is incorporated, the improvements become even more pronounced, particularly with a generation length of 1024.   For instance, combining Learn2PD and EoTP results in a throughput increase of 22.58× (on GSM8K, 5-shot) and 17.16× (on MBPP, 3-shot) relative to the baseline.   These results demonstrate that our methods are not only effective individually but also highly orthogonal, resulting in compounded acceleration.   More importantly, these efficiency gains have negligible impact on accuracy.   The performance scores of our accelerated methods remain within 1–2 points of the baseline, and in some cases, the score is even slightly improved.

\subsection{Compatibility with Key-Value Cache}
We further evaluate the compatibility of our approach with established Key-Value (KV) Cache techniques by integrating both Dual Cache and Prefix Cache strategies \citep{wu2025fast}. Experiments are conducted on GSM8K with a generation length of 1024 tokens. As summarized in Table \ref{kvcache}, the baseline model (Learn2PD \& EoTP) achieves a throughput of 12.26 TPS, a speed-up of 22.58×, and an accuracy score of 79.83. When augmented with the Dual Cache, the system attains substantially higher efficiency, reaching 31.23 TPS and a 57.51× speedup, albeit with a slight decrease in accuracy (74.00). Similarly, incorporating the Prefix Cache also brings noticeable improvements, yielding 14.79 TPS and a 27.23× acceleration while maintaining a competitive score of 77.71. These results confirm that our method is orthogonal to and fully compatible with standard KV caching mechanisms, demonstrating its ability to leverage such strategies to enhance inference efficiency.

\begin{table}[ht]
\vspace{-0.6em}
    \centering
    % \fontsize{8pt}{10pt}\selectfont
    \footnotesize
    \begin{minipage}[t]{0.49\textwidth}
        \centering
        \caption{A comparison of our method with and without KV Cache. The results show a significant performance improvement when augmented with both Dual and Prefix Caches, underscoring that our method is orthogonal to and fully compatible with existing KV caching strategies.}
        \begin{tabular}{c|c|c|c}
            \toprule
            \textbf{Methods} & \textbf{TPS} & \textbf{Speed} & \textbf{Score}\\
            \midrule
            Learn2PD \& EoTP & $12.26$ & $22.58\times$ & $79.83$ \\
            \midrule
            + Dual Cache & $31.23$ & $57.51\times$ & $74.00$ \\
            \midrule
            + Prefix Cache & $14.79$ & $27.23\times$ & $77.71$\\
            \bottomrule
        \end{tabular}
        \label{kvcache}
    \end{minipage}
    \hfill
    \begin{minipage}[t]{0.49\textwidth}
        \centering
        \caption{A comparison of the acceleration performance using filter models of varying complexity (represented by the number of MLP layers). The results indicate that a two-layer MLP model achieves the optimal balance by providing significant speedup.}
        \begin{tabular}{c|c|c|c}
            \toprule
            \textbf{\# Layers} & \textbf{TPS} & \textbf{Speed} & \textbf{Score}\\
            \midrule
            Single-layer & ${8.77}$ & $2.57\times$ & ${78.62}$\\
            \midrule
            Two-layer & ${14.07}$ & $4.13\times$ & ${78.62}$ \\
            \midrule
            Four-layer & ${11.41}$ & $3.35 \times$ & ${78.85}$ \\
            \bottomrule
        \end{tabular}
        \label{complexity}
    \end{minipage}
    \vspace{-1em}
\end{table}

\subsection{Analysis: Ablation Study}

\paragraph{Effect of Filter Model $f_\theta$ Complexity.} To investigate the impact of the filter model’s architectural complexity on acceleration performance, we conduct an ablation study using MLP-based filter models with varying depths. As illustrated in Table \ref{complexity}, a two-layer MLP achieves the highest throughput (14.07 TPS) and speed-up (4.13×), while maintaining 78.62\% accuracy. In comparison, the single-layer model yields lower efficiency with a similar accuracy, suggesting limited representational capacity. Although the four-layer model attains a marginally better accuracy score, it results in reduced inference speed, indicating increased computational overhead. These results demonstrate that a two-layer configuration offers the optimal trade-off between efficiency and predictive performance, effectively balancing model complexity and acceleration gain.

\begin{wraptable}{r}{0.49\textwidth}
    \vspace{-1.6em}
    \scriptsize
    \centering
    \caption{Performance comparison of our methods across different generation lengths. While maintaining a comparable accuracy, both Learn2PD and EoTP deliver substantially greater speedup at a length of 1024 compared to shorter sequences.}
    \begin{tabular}{c|c|c|c}
        \toprule
        \textbf{Gen Length} & \textbf{Methods} & \textbf{Speed} & \textbf{Score}\\
        \midrule
        \multirow{2}{*}{128} & {Learn2PD} & ${3.29 \times}$ & ${73.92}$\\
        \cmidrule(lr){2-4}
        & {Learn2PD \& EoTP} & ${3.36 \times}$ & ${74.07}$\\
        \midrule
        \multirow{2}{*}{256} & {Learn2PD} & ${4.13 \times}$  & ${78.62}$\\
        \cmidrule(lr){2-4}
        & {Learn2PD \& EoTP} & ${4.21 \times}$ & ${78.62}$\\
        \midrule
        \multirow{2}{*}{512} & {Learn2PD} & ${6.66 \times}$  & ${77.71}$\\
        \cmidrule(lr){2-4}
        & {Learn2PD \& EoTP} & ${7.60 \times}$ & ${79.68}$\\
        \midrule
        \multirow{2}{*}{1024} & {Learn2PD} & ${12.21 \times}$  & ${77.26}$\\
        \cmidrule(lr){2-4}
        & {Learn2PD \& EoTP} & ${22.58 \times}$ & ${79.83}$\\
        \bottomrule
        \end{tabular}
    \label{genlength}
  \vspace{-1em}
\end{wraptable}

\paragraph{Effect of Generation Length.} To examine the impact of generation length on the performance of our methods, we compare the speedup and accuracy of Learn2PD and its enhanced variant (Learn2PD \& EoTP) across varying output lengths. As shown in Table \ref{genlength}, both methods achieve greater acceleration as the generation length increases, while consistently maintaining competitive scores. At a shorter length of 128, Learn2PD \& EoTP reaches a speed-up of 3.36×. And the speed-up improves steadily with longer sequences, culminating in a substantial 22.58× acceleration at a length of 1024. These results indicate that our approach is particularly effective for long-sequence generation, efficiently reducing the unnecessary decodings to maximize inference speed without compromising output quality.

\begin{wrapfigure}{r}{0.5\textwidth}
    \vspace{-1.1em}
    \centering
    \includegraphics[width=1\linewidth]{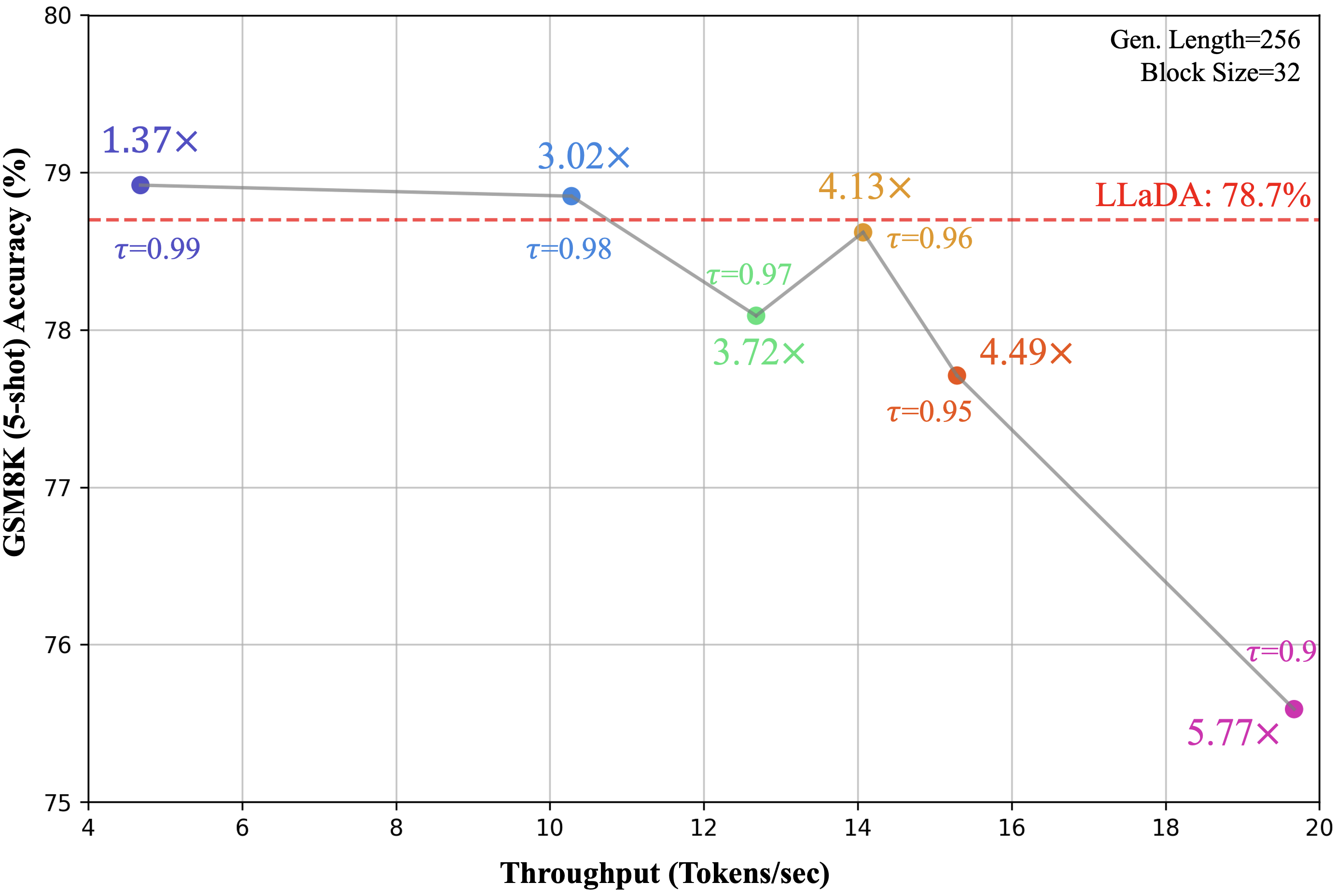}
    \vspace{-0.25in}
    \caption{\textbf{Impact of the filtering threshold on accuracy and throughout.} We find that a threshold of 0.96 represents a favorable balance, maintaining high accuracy and comparable inference speed.}
    \label{fig:threshold}
    % \vspace{-2em}
\end{wrapfigure}

\paragraph{Effect of Filter Model $f_\theta$ threshold $\tau$.} We perform an ablation study to examine the impact of the filtering threshold on inference accuracy and throughput. As shown in Figure \ref{fig:threshold}, reducing the threshold improves throughput but leads to a corresponding decline in accuracy. For example, at $\tau$ = 0.99, the model achieves a throughput of 4.68 TPS (vs. baseline 3.41 TPS) with 78.92\% accuracy. In contrast, lowering the threshold to $\tau$ = 0.9 causes a more pronounced reduction in accuracy. The results indicate that a threshold of $\tau$ = 0.96 offers an optimal balance, delivering both high throughput (4.13× speedup) and near-baseline accuracy. These findings underscore the critical role of the filtering threshold in achieving an effective trade-off between inference efficiency and output quality.

%% file: iclr2026/section/5con.tex
\section{Conclusion}
In this work, we investigate the issue of extensive repetitive decoding during inference in Diffusion-based Large Language Models. To enable timely unmasking of correctly predicted tokens, we propose \textbf{Learn2PD}, a parallel decoding architecture that employs a filter model to make case-specific selections. This filter model is lightweight and pre-trained, thus requiring no additional training during inference. Furthermore, to address the time overhead caused by repeated encoding of the \texttt{[EoT]} token as the generation length increases, we introduce the \textbf{EoTP} mechanism, which halts decoding immediately after \texttt{[EoT]} is generated, thereby reducing unnecessary computational cost. Extensive experiments across multiple benchmarks and model baselines (LLaDA) demonstrate that our approach achieves up to \textbf{22.58×} speedup without sacrificing accuracy—and up to \textbf{57.51×} when combined with KV Cache. Our proposed method offers a compelling solution for deploying diffusion-based LLMs as alternatives to autoregressive models in future applications.

%% file: iclr2026/section/6appendix.tex
% \section{The Use of Large Language Models}
% Large language models (LLMs) were used in this work exclusively for the purpose of text polishing and refinement. Their role was strictly limited to assisting with grammatical correction, improving sentence fluency, and enhancing word choice to increase the overall clarity and readability of the manuscript. 
% All intellectual contributions, including research ideation, methodological design, data analysis, interpretation of findings, and the original drafting of the content, were performed solely by the human authors. 

\section{Update Inference Algorithm}
\label{appendix:update_algorithm}

\begin{algorithm}[H]
\caption{\textbf{Update Inference}}
\begin{algorithmic}[1]
\Require Diffusion large language model $M$, filter model $f_\theta$, prompt set $x_\text{prompt}$, generation length $L_\text{gen}$, block size $s$, filter threshold $\tau$
\For{each $x_i \in x_\text{prompt}$}
    \State $l_i \gets length(x_i)$,  $X \gets \text{concat}(x_i, [\text{MASK}]^{L_\text{gen}})$
    \For{$b = 0,...,\frac{L_\text{gen}}{s}-1$}
        \State $\mathcal{M} \gets \{1,2,...,s\}$
        \While{$\mathcal{M} \neq \emptyset$ }
            \State $conf_t$, $pre_t= M(X)$, $logit_t = f_\theta(conf_t)$
            \If{$logit_{t,j} > \tau$}
                \State $\mathcal{M} \gets \mathcal{M} \setminus \{j\}$, $X_{i+b\cdot s+j} \gets pre_{t,j}$
            \EndIf
        \EndWhile
        \If{\text{[endoftext]} in $X$} \State \textbf{break}
        \EndIf
    \EndFor
    \State $response_i = X_{l_i:l_i+L_\text{gen}-1}$
\EndFor
\State \textbf{return} $response$
\end{algorithmic}
\label{inference-update}
\end{algorithm}

\section{Experiments and Analysis on EoTP mechanism}
\label{appendix:eotp_analysis}
\begin{table}[h]
    % \vspace{-2em}
    \centering
    \caption{A comparison of LLaDA and EoTP mechanisms in different generation lengths.}
    \begin{tabular}{c|c|c|c|c}
        \toprule
        \textbf{Methods} & \textbf{Generation Length} & \textbf{TPS} & \textbf{Speed} & \textbf{Score}\\
        \midrule
        \multirow{3}{*}{LLaDA} & 256 & 3.41 & $1.00\times$ & 78.70\\
        \cmidrule{2-5}
        & 512 & 1.67 & $1.00\times$ & 77.71\\
        \cmidrule{2-5}
        & 1024 & 0.54 & $1.00\times$ & 77.62\\
        \midrule
        \multirow{3}{*}{+ EoTP} & 256 & 3.76 & $1.10\times$ & 79.23\\
        \cmidrule{2-5}
        & 512 & 3.38 & $2.02\times$ & 78.77\\
        \cmidrule{2-5}
        & 1024 & 3.25 & $5.98\times$ & 79.38\\
        \bottomrule
    \end{tabular}
    \label{tab:eotp}
\end{table}

This table \ref{tab:eotp} clearly demonstrates the significant advantage of integrating the EoTP mechanism with the base LLaDA model, particularly for long-sequence generation. While the standalone LLaDA model exhibits substantial performance degradation as generation length increases—evidenced by the sharp decline in TPS from 3.41 to 0.54—the incorporation of EoTP not only mitigates this degradation but also delivers considerable speedup. Most notably, at a sequence length of 1024, EoTP achieves a dramatic 5.98× acceleration while simultaneously improving output quality, with the Score increasing from 77.62 to 79.38. These gains can be largely attributed to the elimination of redundant computation: quantitative analysis shows that 89.59\% of the baseline computational cost arises from decoding padding tokens after the \textbf{\texttt{[EoT]}} token. EoTP effectively removes this overhead by dynamically detecting sequence completion once all original non-\texttt{[MASK]} positions are unmasked, thereby optimizing inference efficiency without compromising performance.

\section{Training Curve of Filter model}
\label{appendix:training_curve}

\begin{figure}[h]
\centering
% \vspace{-0.2in}
\includegraphics[width=0.8\textwidth]{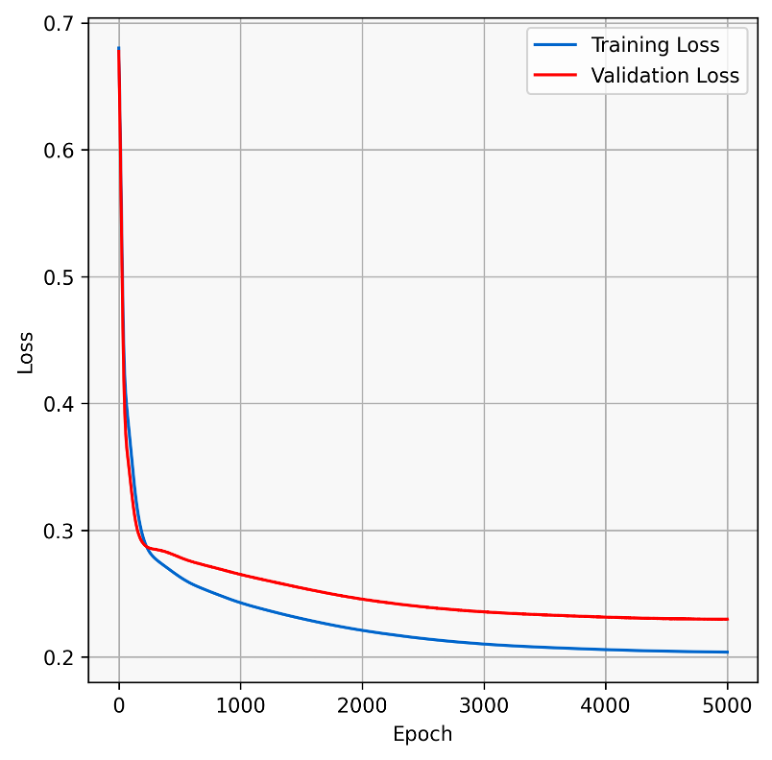}
% \vspace{-0.25in}
\caption{\textbf{Learning curve of filter model $f_\theta$.} The learning curves illustrate the progression of training and validation loss across 5,000 epochs, with the former in blue and the latter in red.} 
% \vspace{-0.2in}
\label{learning_curves}
\end{figure}

As shown in Figure \ref{learning_curves}, this learning curve illustrates that both training (blue) and validation (red) loss fall sharply in the earliest epochs (from 0.68 to 0.28), then decrease much more gradually over the full 5,000-epoch run, approaching a plateau near 0.21–0.23. The validation curve remains slightly above the training curve throughout, indicating only a modest generalization gap rather than pronounced overfitting; the parallel, steady decline shows the model continues to improve on unseen data but with diminishing returns. The long, flat tail of both curves indicates the model has effectively converged.

